\documentclass[runningheads,a4paper]{llncs}

\usepackage{latexsym}
\usepackage{amssymb}
\usepackage{graphicx}
\usepackage{amsmath}
\usepackage{eqnarray}
\usepackage{subfigure}
\usepackage{setspace,cite}
\usepackage{hyperref}
\usepackage{mathtools}
\usepackage{mdwlist}

\hypersetup{colorlinks,citecolor=red,filecolor=black,linkcolor=blue,urlcolor=black}

\usepackage{url}

\urldef{\mailsa}\path|{darshan.venkatrayappa, philippe.montesinos, daniel.diep, baptiste.magnier}@mines-ales.fr|

\begin{document}

\mainmatter  

\title{RSD-DOG : A New Image Descriptor based on Second Order Derivatives}

\titlerunning{RSD-DOG}

%
%
\author{Darshan Venkatrayappa \and Philippe Montesinos\and Daniel Diep\and Baptiste Magnier }

%
\authorrunning{Venkatrayappa et al.}

\institute{LGI2P - Ecole des Mines d'Ales,France\\
\mailsa\\
}

%
%

\toctitle{Lecture Notes in Computer Science}
\tocauthor{Authors' Instructions}
\maketitle

\begin{abstract}
\emph{This paper introduces the new and powerful image patch descriptor based on second order image statistics/derivatives. Here, the image patch is treated as a  3D surface with intensity being the 3rd dimension. The considered 3D surface has a rich set of second order features/statistics such as ridges, valleys, cliffs and so on, that can be easily captured by using the difference of rotating semi Gaussian filters. The originality of this method is based on successfully combining the response of the directional filters with that of the Difference of Gaussian (DOG) approach. The obtained descriptor shows a good discriminative power when dealing with the variations in illumination, scale, rotation, blur, viewpoint and compression. The experiments on image matching, demonstrates the advantage of the obtained descriptor when compared to its first order counterparts such as SIFT, DAISY, GLOH, GIST and LIDRIC.\\}

\end{abstract}

%
%
\section{Introduction}
\label{sec:intro}
Local image feature extraction has evolved into one of the hot research topics in the field of computer vision. Extracting features that exhibit high repeatability and distinctiveness against variations in viewpoint, rotation, blur, compression, etc., is the basic requirement for many vision applications such as image matching, image retrieval, object detection, visual tracking and so on. For this purpose, a number of feature detectors \cite{MikolajczykS04} and descriptors \cite{Evaluation} have been proposed. In the computer vision literature, features related to first order image statistics such as segments, edges, image gradients and corners have been used in abundance for image matching and object detection. Whereas, features related to second order statistics such as cliff, ridges, summits, valleys and so on have been sparsely used for the image matching and object recognition purpose. The scope of the work lies in the use of second order statistics for the task of image matching.

 \begin{figure}
\begin{center}
\begin{tabular}{ccc}       
    
    \includegraphics[height =3.8cm]{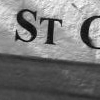}&
    \includegraphics[height =3.8cm]{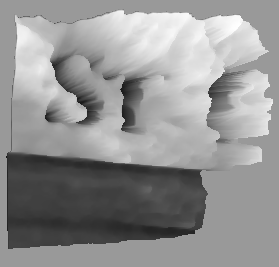}&
    \includegraphics[height =3.8cm]{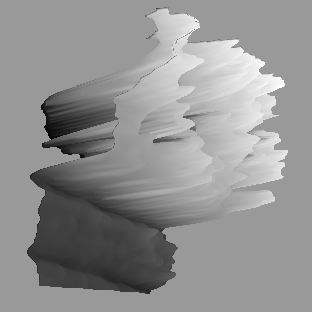} \\
   (a) & (b) &(c) 
%
\end{tabular}
\end{center}
\caption{(a) Image representation in 2D. (b) and (c) The same image being represented in 3D, with intensity being the third dimension. Both (b) and (c) when viewed from different angles, exhibits a set of second order statistics such as ridges, valleys, summits, edges etc. }
\label{3D-representation}
\end{figure}

\subsection{Related work}
   In one dimension, the first order gradient extracted at a point gives the slope of the curve at the given point. In case of an image, the first order gradient at a pixel measures the slope of the luminance profile at that pixel. 
Several local image descriptors such as SIFT \cite{SIFT}, GLOH \cite{Evaluation}, DAISY \cite{DAISY}, and LBP \cite{LBP} are based on first order gradient information of the image. And amongst all, SIFT is one of the most widely used local image descriptor. It is constructed by capturing the gradient information at every pixel around an interest point. Extensions of SIFT such as PCA-SIFT \cite{PCA-SIFT}, F-SIFT \cite{F-SIFT}, MI-SIFT \cite{MI-SIFT} are an improved version of the original SHIFT descriptor, by introducing the new invariance properties while using the same first order gradients as their bases. While GLOH \cite{Evaluation} improves on the robustness and distinctiveness of the SIFT descriptor by using radial binning strategy, Fan et al. \cite{MRGOH} pool the first order image gradients based on their intensity orders in multiple support regions. By doing so, they achieve rotation invariance without actually calculating the reference orientation. DAISY, combines both SIFT and GLOH binning strategy for fast and dense matching.

    Ojala et al. \cite{LBP} came up with local descriptor made of first order binary patterns (LBP) for texture classification. Center-Symmetric LBP (CS-LBP) \cite{CS-LBP} and orthogonal color LBP (OC-LBP) \cite{OC-LBP} provides a compact representation of the LBP descriptor while keeping the same discriminative power. Zambanini et al. \cite{LIDRIC} propose LIDRIC descriptor, based on multi-scale and multi-oriented even Gabor filters. The descriptor is constructed in such a way that typical effects of illumination variations like changes of edge polarity or spatially varying brightness changes at each pixel are taken into account for illumination insensitivity. LIDRIC has a dimension of 768. Oliva et al. \cite{GIST} employ Gabor filters to the grey-scale input image at four different angles and at four spatial scales to obtain the GIST descriptor. The descriptor has a dimension of 512 and is more global. Authors of \cite{RSD-HOG} propose a new descriptor called RSD-HoG. It is based on the orientation of the edges. The edge orientations are extracted by using a rotating half Gaussian kernel. The descriptor is constructed in a novel way, by embedding the response of the rotating half Gaussian kernel in a Histogram of oriented Gradient (HoG) framework.           
   
   On the other hand, in one dimension second order derivative at a point measures the local curvature at that point i.e. how much the curve bends at that point. The authors of \cite{PAMI} use an oriented second derivative filter of Gaussians to capture isotropic as well as anisotropic characteristics of surface by the use of single scale for descriptor generation. 
Fischer et al. \cite{FISHER} proposed a new image descriptor based on second order statistics for image classification and object detection. Their descriptor extracts the direction and magnitude of the curvature and embeds these information in the HOG framework. But, one of the disadvantage of this descriptor is the dimension. Eigenstetter et al. \cite{Eigenstetter} have proposed an object representation framework based on curvature self-similarity. This method goes beyond the popular approximation of objects using straight lines. However, like most of the descriptors using second order statistics, this approach also exhibits a very high dimensionality.
   
   As shown in Fig.\ref{3D-representation}, An image represented by a two dimensional function can be considered as a surface in a 3D space. Such a surface in 3D, consists of features such as ridges, valleys, summits or basins. The geometric properties of these features can be accurately characterized by local curvatures of differential geometry through second order statistics. The motivation behind this work is to extract these 2nd order image statistics and represent them as a compact image descriptor for image matching. The main contribution of this work is:
\begin{enumerate}
\item The idea was to consider the 2D image patch as 3D surface made of ridges, valleys, summits, etc and to extract these second order statistics by using a local directional maximization or  minimization of the response of difference of two rotating half smoothing filters.
\item These directions correspond to the orientation of ridges, valleys or a junction of ridges/valleys. The orientations at which these second order statistics occur, are binned to form a local image descriptor RSD-DOG of dimension/length 256. By construction, the dimension of the descriptor is almost 3 to 4 times less when compared to other descriptors based on second order statistics.
\item This descriptor is evaluated for invariance to blur, rotation, compression, scale and viewpoint changes. Additionally, by construction, the descriptor shows enormous robustness to variations in illumination. To highlight this property, we rigorously evaluate the descriptor on dataset consisting of images with linear and non-linear illumination changes.
\end{enumerate}
The remainder of the paper is organized as follows. In the section 2, we present a directional filter made of anisotropic
smoothing half Gaussian kernels. In section 3, we present our robust method for extracting the ridge/valley directions using difference of half directional Gaussian filters. Section 4 is devoted to descriptor construction process and section 5, discusses about experiments and results. Finally, section 6 concludes this paper with the future work.

\section{Directional Filter}
\begin{figure}[t!]
\begin{center}
\begin{tabular}{cccc}
 \includegraphics[height=2.5cm]{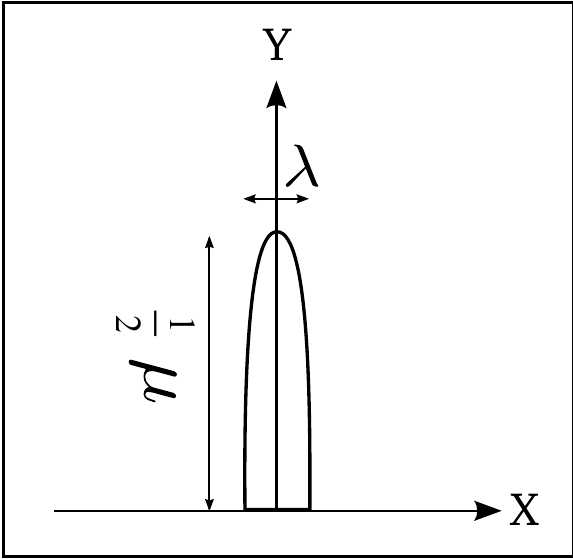} &
   \includegraphics[height=2.5cm]{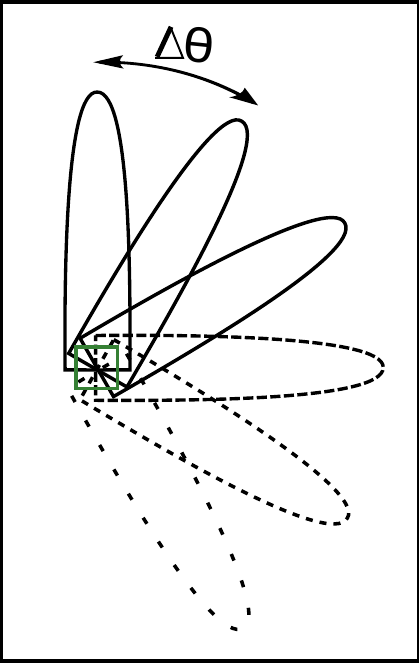} &
  \includegraphics[height=2.5cm]{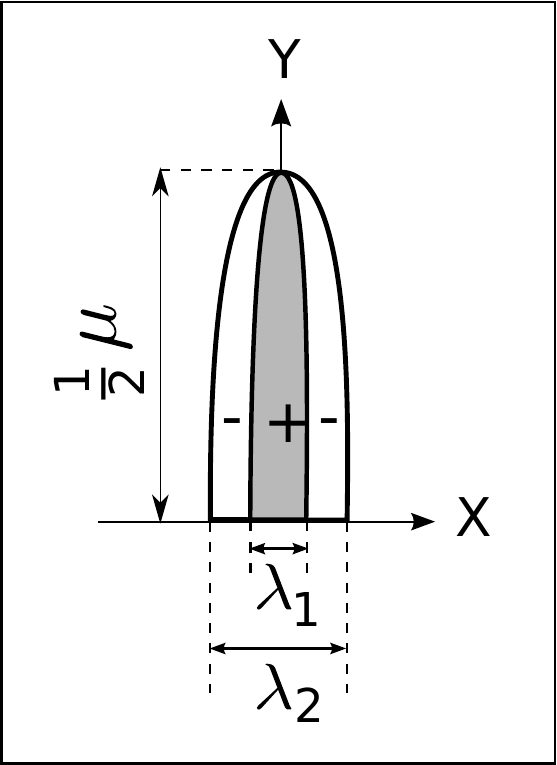} &
  \includegraphics[height =2.6cm]{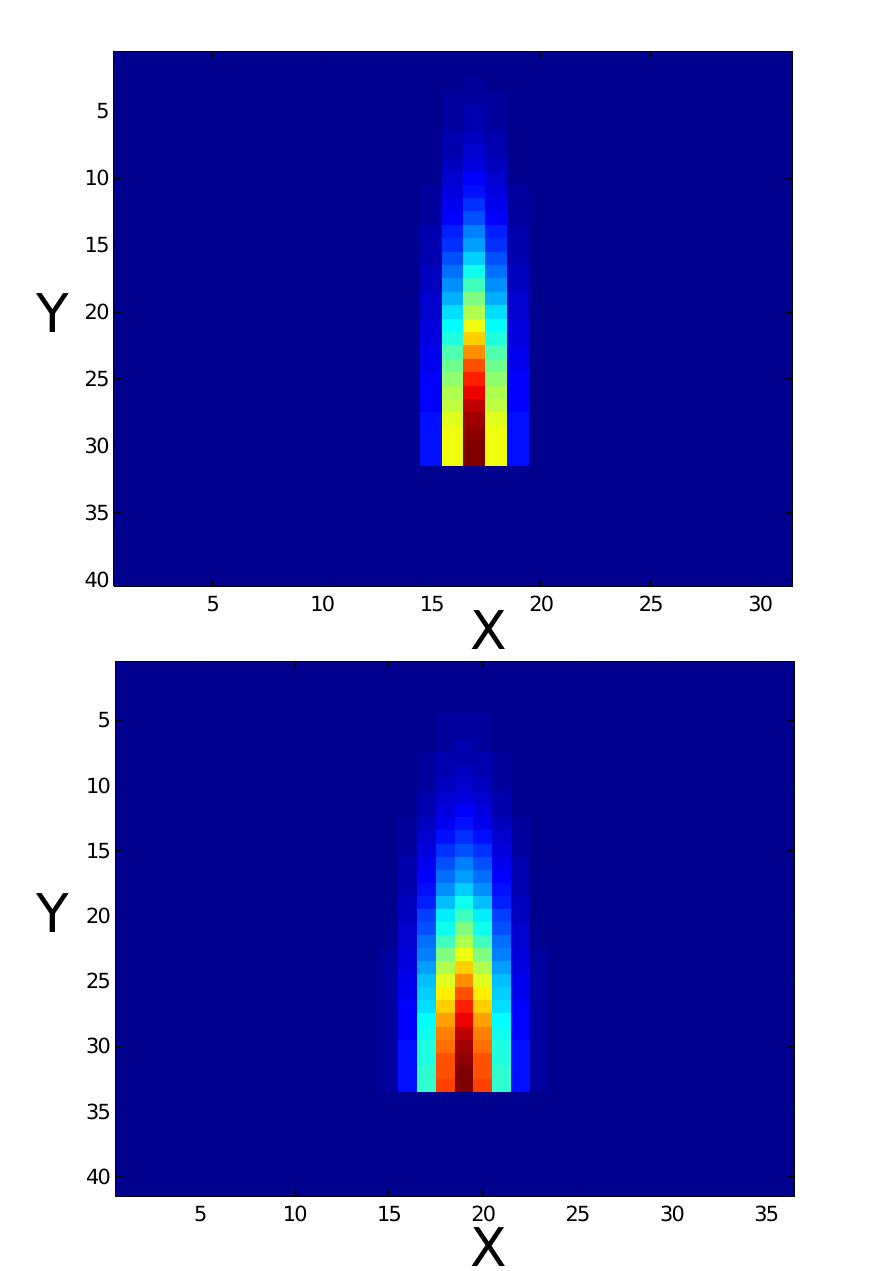} \\
   \vspace{0.2cm}
(a) \scriptsize{Smoothing filter} & (b) \scriptsize{Rotating filters} & 
(c) \scriptsize{DHSF.} &
(d) \scriptsize{Discretized DHSF.}\\
\end{tabular}
\end{center}
\vspace{-0.5cm}
\caption{A smoothing rotating filter and Difference of Half smoothing Filters (DHSF). For (d) top: $\mu = 10$ and $\lambda = 1$. For (d) bottom:  $\mu = 10$ and $\lambda = 1.5$.}
\label{croix_figure}
\end{figure}

\noindent In our method, we use a directional filter made of anisotropic smoothing half Gaussian kernels. For every pixel in the image patch, we spin this directional filter to obtain a Rotating Signal Descriptor (RSD), which is a function of a rotation angle $\theta$ and the underlying signal.  As shown in \cite{Montesin} and \cite{Magnier2011}, smoothing with rotating filters means that the image patch is smoothed with a bank of rotated anisotropic Gaussian  kernels:

 {
\begin{eqnarray}
G_{(\mu,\lambda)}(x,y, \theta)=C. H \left( R_\theta 
                         \left(\begin{array}{l}x \\
                                               y \end{array}\right)\,\right) e^{ -\left(\begin{array}{ll} x & y \end{array} \right)
                         R^{-1}_\theta 
                         \left(\begin{array}{ll}\frac{1}{2\ \mu^2} &        0 \\
                                                                 0 & \frac{1}{2 \lambda^2}
                         \end{array} \right)
                         R_\theta 
                         \left(\begin{array}{l}x \\
                                               y \end{array}\right)
                   }
 \label{gaussian_eq}
 \end{eqnarray}
 }
where $C$ is a normalization coefficient, $R_\theta$ a rotation matrix of angle $\theta$, $x$ and $y$ are pixel coordinates and $\mu$ and $\lambda$ are the standard-deviations of the Gaussian filter. As we require only the causal part of the filter (illustrated on figure \ref{croix_figure}(a)), we simply ``cut'' the smoothing kernel at the middle, and this operation corresponds to the Heaviside function $H$ \cite{Montesin}. By convolving the image patch with these rotated kernels (see figure \ref{croix_figure}(b)), we obtain a stack of directional smoothed image patches $I_\theta = I \ast G_{(\mu,\lambda)}(\theta)$. 

To reduce the computational complexity, in the first step we rotate the image at some discrete orientations from 0 to 360 degrees (of $\Delta\theta =$ 1, 2, 5, or 10 degrees, depending on the angular precision required and the smoothing parameters)
before applying non rotated smoothing filters. 
As the image is rotated instead of the filters, the filtering implementation can use efficient recursive approximation of the Gaussian filter. As presented in \cite{Montesin}, the implementation of the method is clear and direct. In the second step, we apply an inverse rotation of the smoothed image and obtain a bank of $360/ \Delta\theta$ images.

\section{Ridge and valley detection using difference of Gaussian filters}
\subsection{Difference of Half smoothing filters (DHSF)}
\begin{figure}
\begin{center}
\begin{tabular}{cc}       \vspace{0.2cm}
    \includegraphics[height=3cm]{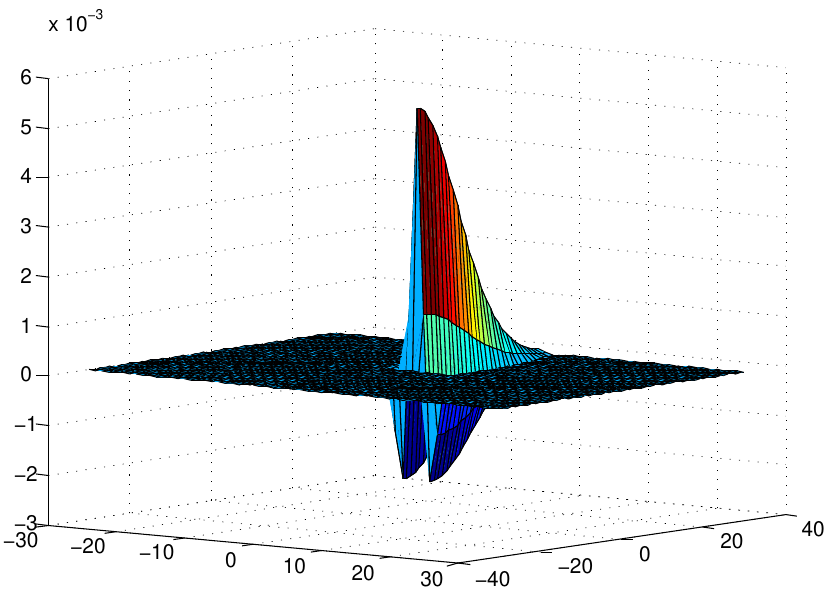} &
  \includegraphics[height =3cm]{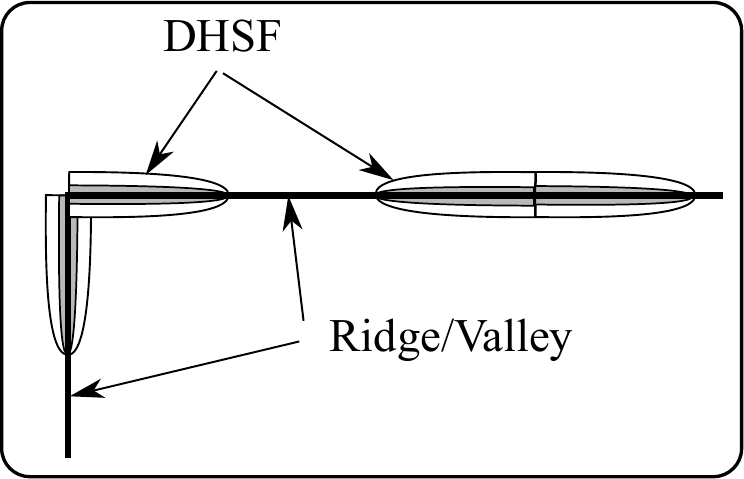} \\

(a) \small{A DHSF in 3 dimension} & 
(b) \small DHSF in the ridge/valley directions \\
\end{tabular}
\end{center}
\vspace{-0.5cm}
\caption{DHSF filter descriptions.}
\label{gaussian_dog_rot}
\end{figure}

At every pixel in the image patch, we are required to estimate a smoothed second order derivative of the image along a curve crossing these pixels. In one dimension, the second order derivative of a signal can be easily estimated using a Difference Of Gaussian (DOG)operator \cite{SIFT}. In our method, as in \cite{ACIVS2011}, we directly apply two half Gaussian filters with two different $\lambda$ and the same $\mu$ to obtain the directional derivatives. An example for the two discretized filters is shown in Fig.\ref{croix_figure}(d). Later, we compute the difference of the response of these two filters to obtain the desired smoothed second order derivative information in the ridge/valley directions (illustrated in Fig. \ref{gaussian_dog_rot}(b)). We refer to this half Gaussian filter combination as the difference of half smoothing filters (DHSF). An illustration of DHSF is presented in Fig.\ref{gaussian_dog_rot}(a) and Fig.\ref{croix_figure}(c).


\subsection{Estimating the direction of second order statistics such as ridges and valleys}
 By convolving the image patch with the DHSF (by the technique of rotating the images, as explained above), for each pixel in the image patch we obtain a pixel signal which captures a directional second order information around the pixel. The idea is to extract the directions at which these second order statistics such as ridges and valleys occur and to construct a descriptor from this information. Let us consider $D(x,y,\theta)$ to be the pixel signal obtained at pixel P located at $(x,y)$.  $D(x,y,\theta)$ is a function of the direction $\theta$ such that:

 {
\begin{eqnarray}
D(x,y, \theta) = G_{(\mu,\lambda_1)}(x,y, \theta) - G_{(\mu,\lambda_2)}(x,y,\theta)
 \label{gaussian_diff_eq}
 \end{eqnarray}
 }
 $\mu$, $\lambda_1$ and $\lambda_2$ correspond to the standard-deviations of the Gaussians. At each pixel in the image patch, we are interested in the response of the DHSF at $\theta_{M_1}$, $\theta_{M_2}$, $\theta_{m_1}$ and $\theta_{m_2}$. Where, $\theta_{M_1}$ and $\theta_{M_2}$ are the directions at which the local maxima of the function $D$ occurs. $D(x,y,\theta_{M_1})$ and $D(x,y,\theta_{M_2})$ are the response of DHSF at $\theta_{M_1}$ and $\theta_{M_2}$. $\theta_{m_1}$ and $\theta_{m_2}$ are the directions at which the local minima of the function $D$ occurs. $D(x,y,\theta_{m_1})$ and $D(x,y,\theta_{m_2})$ are the response of DHSF at $\theta_{m_1}$ and $\theta_{m_2}$.
 
\begin{figure}[t]
  \centering
  \includegraphics[height=1.4cm, angle=90]{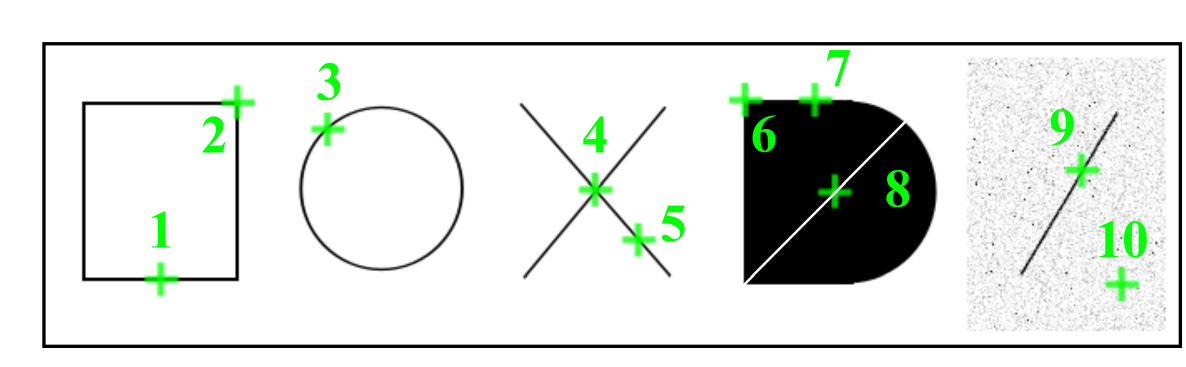}
   \includegraphics[height=4.8cm]{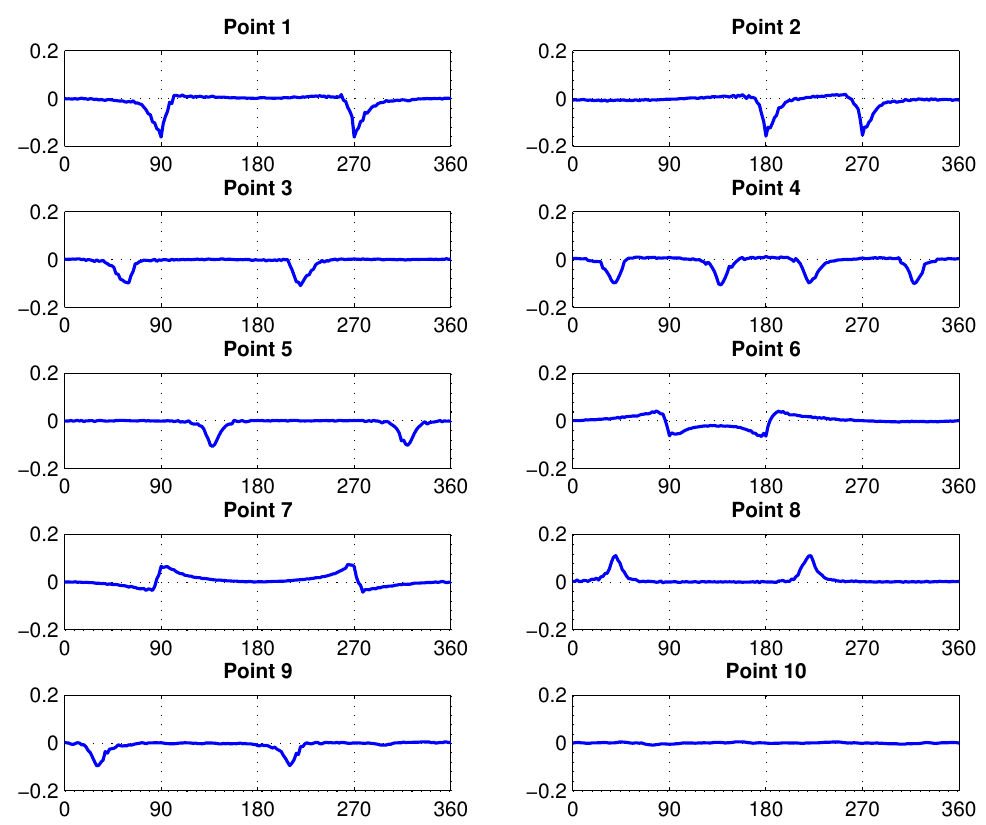}
\vspace{-.3cm}
\caption{Points selection on a synthetic image. Examples of functions $D(x,y,\theta)$ on the points selected on synthetic image using $\mu = 10, \lambda_1 = 1, \lambda_2 = 1.5$. \it{The x-axis corresponds to the value of $\theta$ (in degrees) and the y-axis to $D(x,y,\theta)$}. }
\label{signaux_points}
\end{figure}

 Some examples of the signal $D(x,y,\theta)$ obtained by spinning the DHSF around the selected key-points extracted from the synthetic image are shown in Fig.\ref{signaux_points}. On a typical valley (point 1 in Fig. \ref{signaux_points}), the pixel signal at the minimum of a valley consists of at least two negative sharp peaks. For ridges (point 7 in Fig. \ref{signaux_points}), the pixel signal at the maximum of a ridge contains at least two positive peaks. These sharp peaks correspond to the two directions of the curve (an entering and leaving path). In case of a junction, the number of peaks corresponds to the number of crest lines (ridges/valleys) in the junction (point 4 in Fig. \ref{signaux_points}).
We obtain the same information for the bent lines (illustrated in point 2 on Fig. \ref{signaux_points}).
Finally, due to the strong smoothing (parameter $\mu$), $D$ is close to $0$ in the presence of noise without any crest line nor edge (illustrated in point 10 in Fig. \ref{signaux_points}). This illustrates the robustness of this method in the presence of noise.


\section{DESCRIPTOR CONSTRUCTION}

\begin{figure}[b!]
\centering
 \includegraphics[width=11cm]{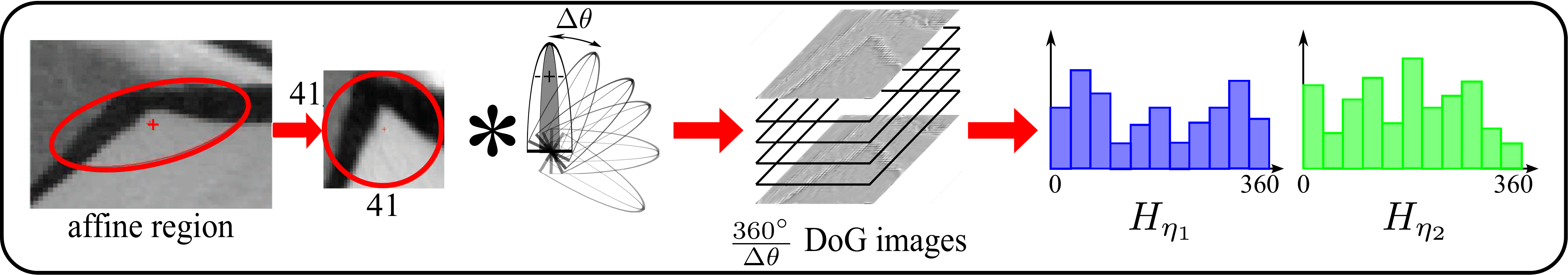}\\
 \vspace{-0.3cm}
 \caption{Methodology involved in the construction of RSD-DOG descriptor.}
\label{descriptor_compute}
\end{figure}
The descriptor construction process as shown in Fig.\ref{descriptor_compute}, in the initial stage, as in \cite{MikolajczykS04}, for each detected key-point we follow the standard procedure to obtain the rotation and affine normalized gray level image patch. This normalization procedure is followed in the construction of all the descriptors (SIFT, DAISY, GLOH) used in our experiments. We consider this image patch as a 3D surface, with intensity being the 3rd dimension. As in Fig.\ref{descriptor_compute}, for each pixel in the image patch, we spin the DHSF and obtain a stack of DOG patches. From this stack of DOG patches, for each pixel we extract the signal $D(x,y,\theta)$ (for simplicity and proper viewing, in Fig.\ref{descriptor_compute} signal is not shown and a stack of image patch is shown). From each signal we extract the four angles $\theta_{M_1}$, $\theta_{M_2}$, $\theta_{m_1}$, $\theta_{m_2}$ and their corresponding responses $||D(x,y,\theta_{M_1})||$, $||D(x,y,\theta_{M_2})||$, $||D(x,y,\theta_{m_1})||$ and $||D(x,y,\theta_{m_2})||$. Once these informations are obtained, for each pixel P, we estimate the average angles $\eta_1$ and $\eta_2$ and there respective average magnitudes $\delta_{1}$ and $\delta_{2}$ by:

{
$$\left\{ \begin{array}{ll}
\eta_{1}(x,y) = (\theta_{M_1} + \theta_{M_2}) / 2 \\
\eta_{2}(x,y) = (\theta_{m_1} + \theta_{m_2}) / 2 \\
\delta_{1} = (||D(x,y,\theta_{M_1})|| + ||D(x,y,\theta_{M_2})||) / 2 \\
\delta_{2} = (||D(x,y,\theta_{m_1})|| + ||D(x,y,\theta_{m_2})||) / 2 \\
\end{array}
\right.
$$
}

\begin{figure}[!t]
  \centering
\begin{tabular}{cc}
 \includegraphics[height=3.4cm]{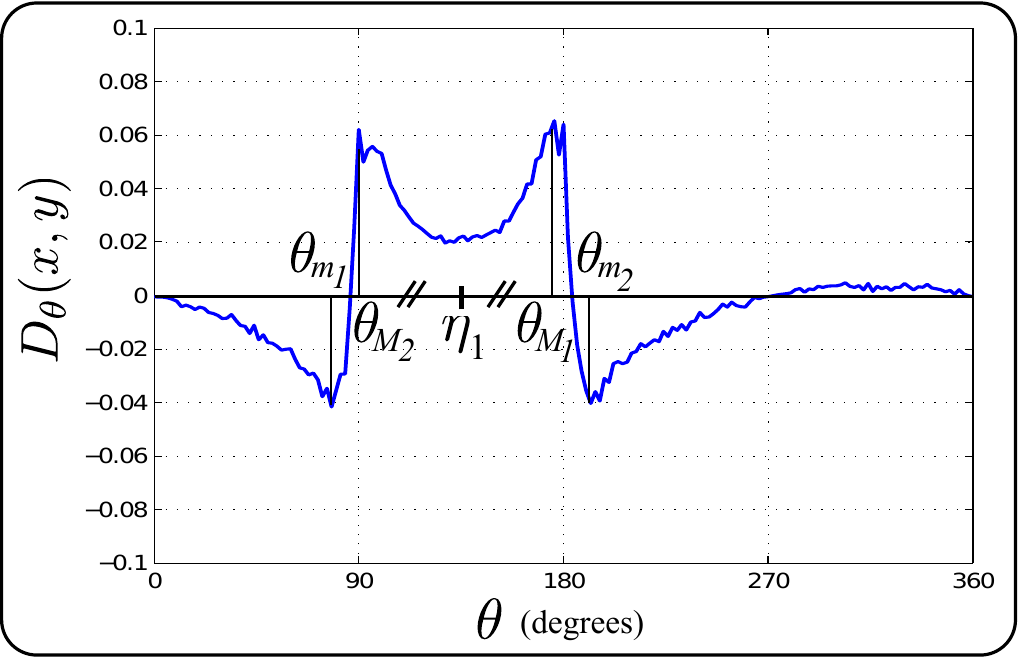} &
   \includegraphics[height=3.4cm]{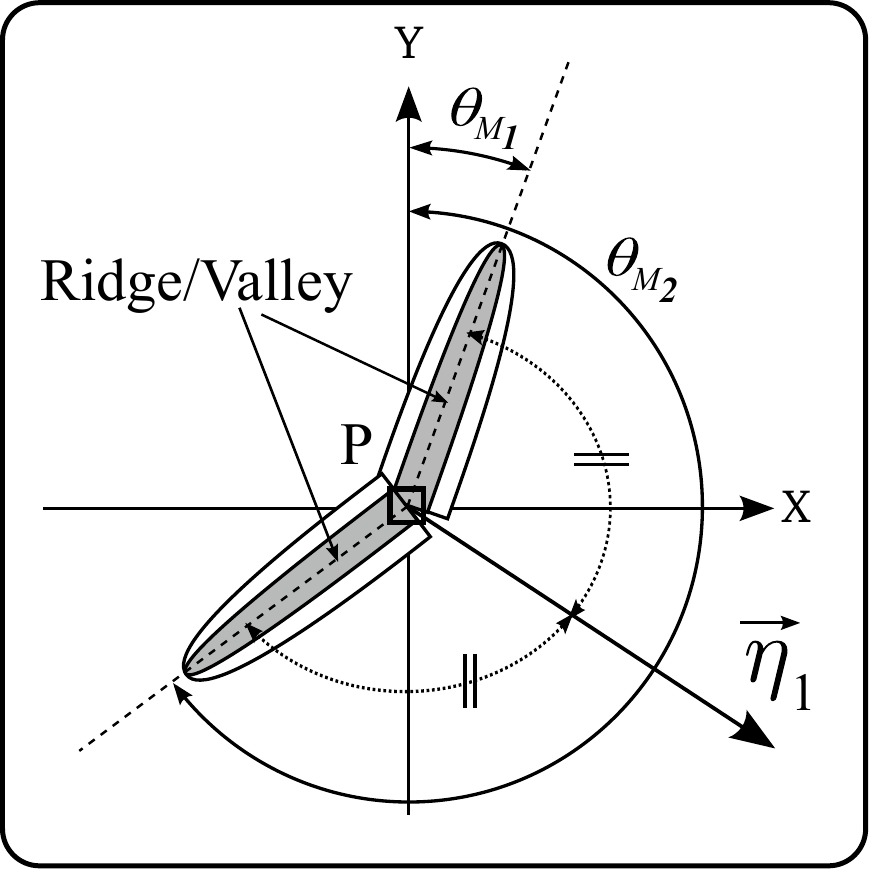} \\
 (a) & (b)\\
\end{tabular}
\vspace{-0.3cm}
\caption{(a) $\eta_1$ computation from $\theta_{M_1}$ and $\theta_{M_2}$. (b) $\eta_1$ corresponds to the direction perpendicular to the ridge/valley at the level of a pixel P.}
\label{eta_compute}
\end{figure}

The angle $\eta_{1}$ is weighed by $\delta_{1}$ and $\eta_{2}$ by $\delta_{2}$ and binned as in Eq. \ref{theta_1_2_3}. Later, $H_{\eta_1}$ and $H_{\eta_2}$ are concatenated to form the final 256 length/dimension RSD-DOG descriptor.

 \begin{equation}
	\label{theta_1_2_3}
	\left \{ 
	\begin{array}{clll}
	H_{\eta_1} = \lbrace\eta_{1_{bin1}}, \eta_{1_{bin2}}, \eta_{1_{bin3}}, \eta_{1_{bin4}}.........\eta_{1_{bin128}} \rbrace\\
	H_{\eta_2} = \lbrace\eta_{2_{bin1}}, \eta_{2_{bin2}}, \eta_{2_{bin3}}, \eta_{2_{bin4}}.........\eta_{2_{bin128}} \rbrace
    \end{array}
	\right .
\end{equation}


\section{Experiments and Results}   
\subsection{Dataset and Evaluation}
 
  Matlab platform is used for the experiments. Harris affine key points \cite{MikolajczykS04} were used for image patch extraction as well as the key points obtained from other detectors can also be used for extracting these image patches. We evaluate and compare the performance of our descriptor as against the state of the art descriptors on the standard dataset, using the standard protocol provided by the Oxford group. The binaries and dataset are publicly available on-line\footnote[1]{http://www.robots.ox.ac.uk/~vgg/research/affine/}. 
 
\begin{figure}[t!]
 \begin{tabular}{cccccccc}

\includegraphics[width=14mm]{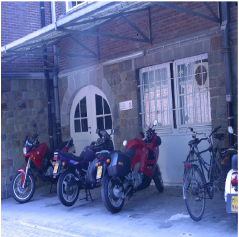}&
\includegraphics[width=14mm]{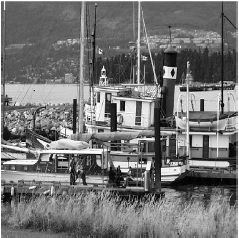}&
\includegraphics[width=14mm]{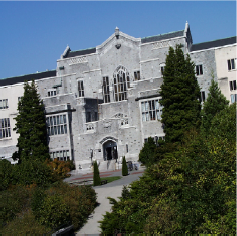}&
\includegraphics[width=14mm]{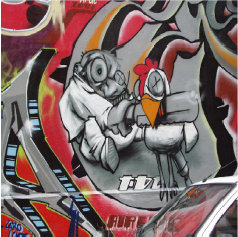}&
\includegraphics[width=14mm]{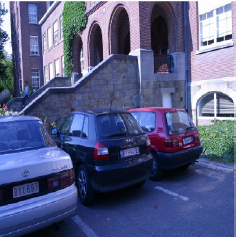}&
\includegraphics[width=14mm]{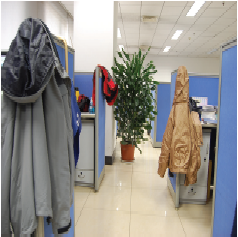}&
\includegraphics[width=14mm]{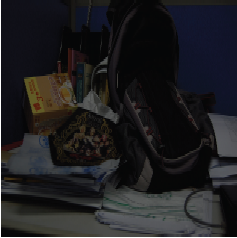}&
 \includegraphics[width=14mm]{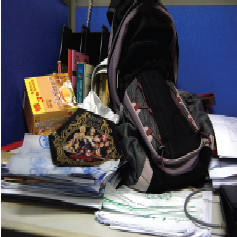} \\ 
\includegraphics[width=14mm]{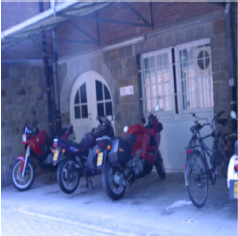}&
\includegraphics[width=14mm]{data/dataset/BOAT2.png}&
\includegraphics[width=14mm]{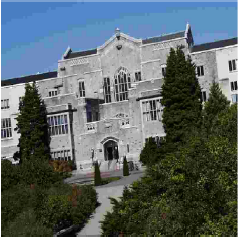}&
\includegraphics[width=14mm]{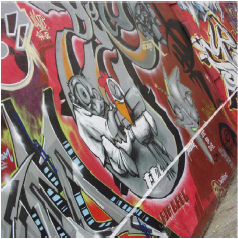}&
\includegraphics[width=14mm]{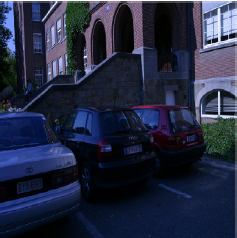}&
\includegraphics[width=14mm]{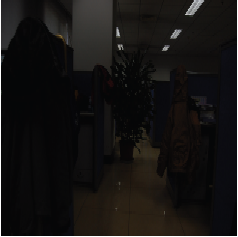}&
\includegraphics[width=14mm]{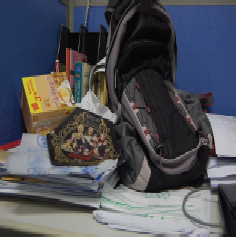}&
\includegraphics[width=14mm]{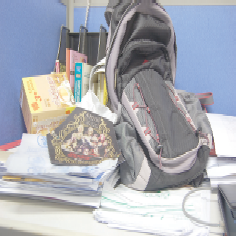} \\ 
\vspace{-.25cm} \\
\scriptsize{(a)}{\label{BIKE-TEST}}&
\scriptsize{(b)}{\label{BOAT-TEST}}&
\scriptsize{(c)}{\label{COMP-TEST}}&
\scriptsize{(d)}{\label{GRAFF-TEST}}& 
\scriptsize{(e)}{\label{LUVEN-TEST}}&
\scriptsize{(f)}{\label{CORRIDOR-TEST}}&
\scriptsize{(g)}{\label{DESKTOP-TEST}}&
\scriptsize{(h)}{\label{SQRT-RT-TEST}} \\

 \end{tabular}
 \vspace{-0.15cm} 
\caption{A few examples of image pairs used in our evaluations. Images in the first column (a) has variations in blur (BIKE), second column (b) has changes in rotation (BOAT), third column (c) has compression changes, fourth column (d) has changes in view-point (GRAFF). Images in the fifth, sixth and seventh column have variations in illumination( (e) LUEVEN, (f) CORRIDOR, (g) DESKTOP, (h) SQUARE-ROOT ). }
\label{DATASET}
\end{figure}

 The dataset used in our experiments has different geometric and photometric transformations, such as, change of scale and image rotation (boat), viewpoint change (graf), image blur (bike), JPEG compression (compression) and illumination change (Leuven). For each type of the image transformation, there is a set of six images with established ground truth homographies. In-order to study in detail the performance of our descriptor for changes in illumination, we also evaluated our descriptors on four image pairs, with complex illumination changes and the data set for the same is publicly available\footnote[2] {http://zhwang.me/publication/liop/}. The complex illumination dataset has 4 set of images, namely 'desktop', 'corridor', 'square' and 'square root'. The first two sets, 'desktop' and 'corridor' have drastic illumination changes, whereas 'square' and 'square root' datasets are obtained by a square and square root operation on the second image of the 'desktop' set \cite{LIOP}. Some of the image pairs from both datasets are shown in Fig.\ref{DATASET}.

The evaluation criterion used as proposed by \cite{Evaluation}, is based on the number of correspondences, correct matches and false matches between two images. Here, we test the descriptors using similarity threshold based matching, since this technique is better suited for representing the distribution of the descriptor in its feature space\cite{Evaluation}. Due to the space limitation, we restrain from going into the details of this method. A detailed description of this method can be found in \cite{Evaluation}. The results are presented using the \textrm{recall vs 1-precision} curves.  As in Eq.\ref{recall}, recall is defined as the total number of correctly matched affine regions over the number of corresponding affine regions between two images of the same scene. From Eq.\ref{precision}, 1-precision is represented by the number of false matches relative to the total number of matches. In all our experiments, Euclidean distance is used as the distance measure.     

  \begin{equation}
    \textrm{recall} = \dfrac{\textrm{Total No of correct matches}}{\textrm{No of correspondences}}
    \label{recall}
  \end{equation}
  \begin{equation}
         \textrm{1-precision} = \dfrac{\textrm{No of false matches} }{\textrm{No of correct matches + No of false matches}}
         \label{precision}
  \end{equation}
  
Our descriptor depends on $5$ different parameters: $\Delta\theta$, $No\!-\!of\!-\!bins$, $\mu$, $\lambda_1$ and $\lambda_2$. The rotation step $\Delta\theta$ is fixed to $10^{\circ}$. Increasing the rotation step results in loss of information. As in \cite{SIFT} the image patch is divided into $16$ blocks. All blocks are of the size 10x10 (Since we are using a patch of size 41x41, the blocks in the extreme right and bottom have 11x11 size). As in \cite{SIFT}, the number of bins ($No\!-\!of\!-\!bins$) is fixed to $8$ per block, resulting in a $8*16=128$ bins for $16$ blocks. Increasing the number of bins results in same performance as in previous case, but it increases the dimensionality of the descriptor. Filter height $\mu$ is fixed to $6$. As in \cite{SIFT}, for DHSF the ratio between successive scales is fixed to $\sqrt{2}$. So, filter widths $\lambda_1$ and $\lambda_2$ are fixed to $2$ and $2\sqrt{2}$ respectively. In our experiments, we obtain state of art results by using just two scales. Height ($\mu = 6$) and  Width ($\lambda_1 = 2$, $\lambda_2 = 2\sqrt{2}$) parameters are chosen to have a ratio sharpness length that is suitable for robust second order feature detection \cite{Montesin}, which generally gives good results in most cases. This ratio is compatible with the angle filtering step.

\subsection{Descriptor Performance}

 In our experiments, we have used two variations of our descriptor (1) RSD-DOG (2-SCALES) with height $\mu=6$ and width $\lambda= 2, 2\sqrt{2}$ respectively. This has a dimension of 256. (2) RSD-DOG (3-SCALES) with height $\mu=6$ and width $\lambda= 2, 2\sqrt{2}, 4$ respectively. Here, in step one, we smooth the image patch with $\mu=6$, $\lambda_1= 2$ and $\lambda_2= 2\sqrt{2}$ to obtain a 256 length descriptor. In the second step, we smooth the image patch with $\mu=6$, $\lambda_1= 2\sqrt{2}$ and $\lambda_2= 4$ and obtain another 256 length descriptor. Lastly, we concatenate the two parts to form a 512 size RSD-DOG(3-SCALES) descriptor.
 
 The performance of these two variants of RSD-DOG is compared with the performance of SIFT, GLOH, DAISY, GIST and LIDRIC descriptors. 
For SIFT and GLOH, the descriptors are extracted from the binaries provided by Oxford group. For DAISY descriptor, the patches are extracted from the code provided by \footnote[3]{http://cvlab.epfl.ch/software/daisy}. 
The matlab code for GIST and LIDRIC descriptors were obtained from \footnote[4]{http://people.csail.mit.edu/torralba/code/spatialenvelope/} and \footnote[5]{http://www.caa.tuwien.ac.at/cvl/project/ilac/} respectively.
 
  For changes in rotation, viewpoint, blur and compression both variants of the RSD-DOG shows better performance than the other 5 descriptors. The precision vs (1-recall) plots in the first 4 rows of the Fig.\ref{EVALUATION} illustrates the superiority of our descriptor. Image pair graf(1-5) is a complex image pair. As a result, performance of  all the descriptors deteriorates. It should be noted that, in most of the cases, RSD-DOG (3-SCALES) performs similar to or slightly better than that of RSD-DOG (2-SCALES). So, increasing the number of scales increases the complexity and descriptor dimension with very little gain in performance. For variations in illumination, in all cases, both the variants of RSD-DOG performs consistently better than all the other descriptors. When it comes to 'square' and 'square root' images SIFT, DAISY, LIDRIC and GIST descriptors exhibit poor performance and GLOH descriptor fails miserably. The graphs in the last 4 rows of Fig.\ref{EVALUATION}, illustrate the superior nature of our descriptor for complex illumination changes. 
        
 
%
%
%
 
\begin{figure}
 \begin{tabular}{ccc}
\includegraphics[height=28mm, width=39mm]{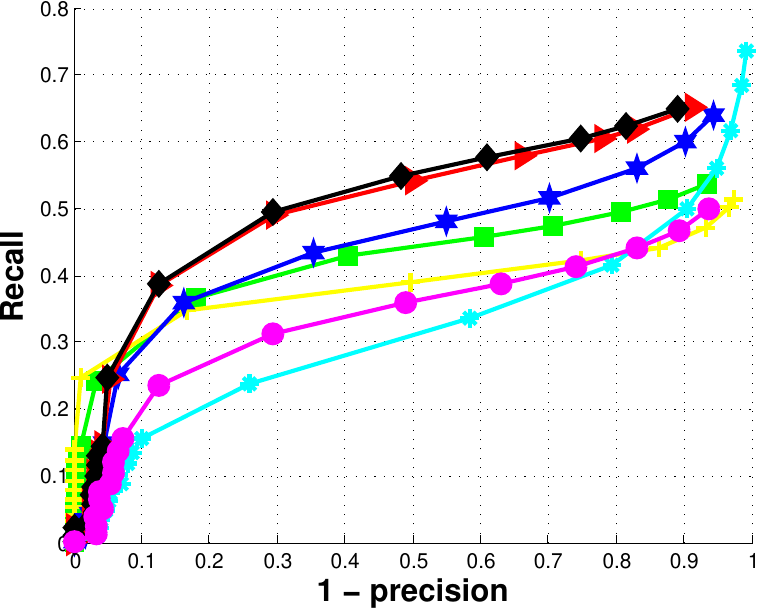}&
\includegraphics[height=28mm, width=39mm]{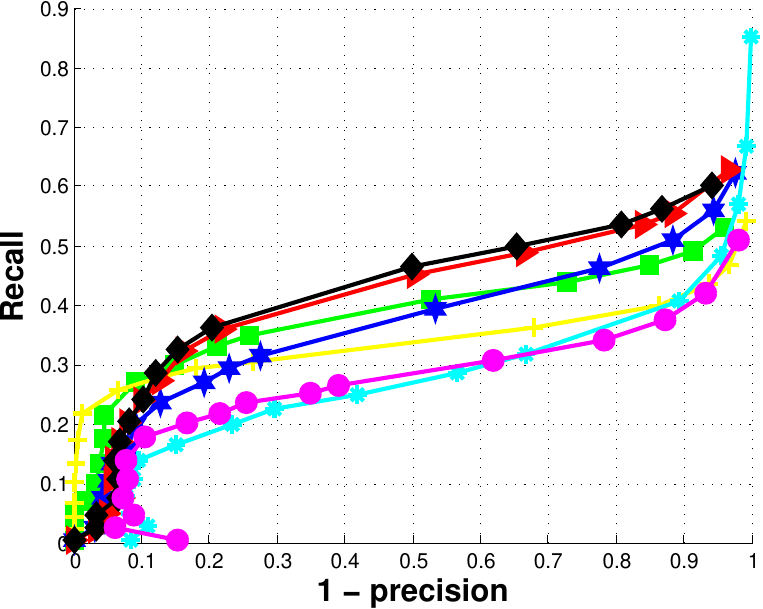}&
\includegraphics[height=28mm, width=39mm]{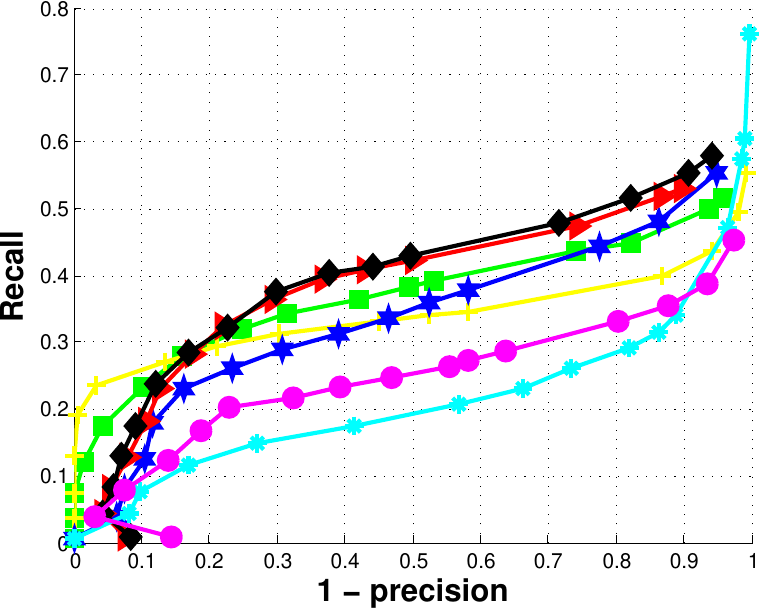}  \vspace{-.25cm} \\
\scriptsize{(b) boat 1-3}{\label{SIMboat13}}&
\scriptsize{(c) boat 1-4}{\label{SIMboat14}}&
\scriptsize{(d) boat 1-5}{\label{SIMboat15}} \\

\includegraphics[height=28mm, width=39mm]{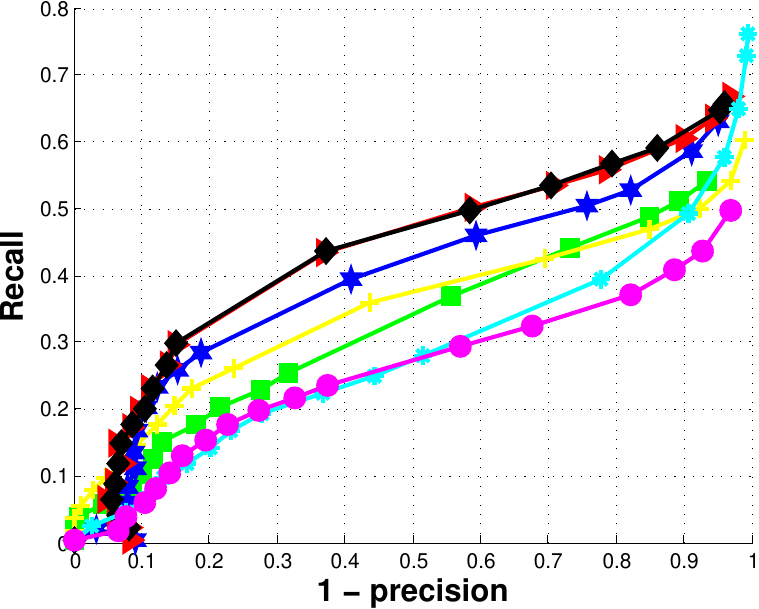}&
\includegraphics[height=28mm, width=39mm]{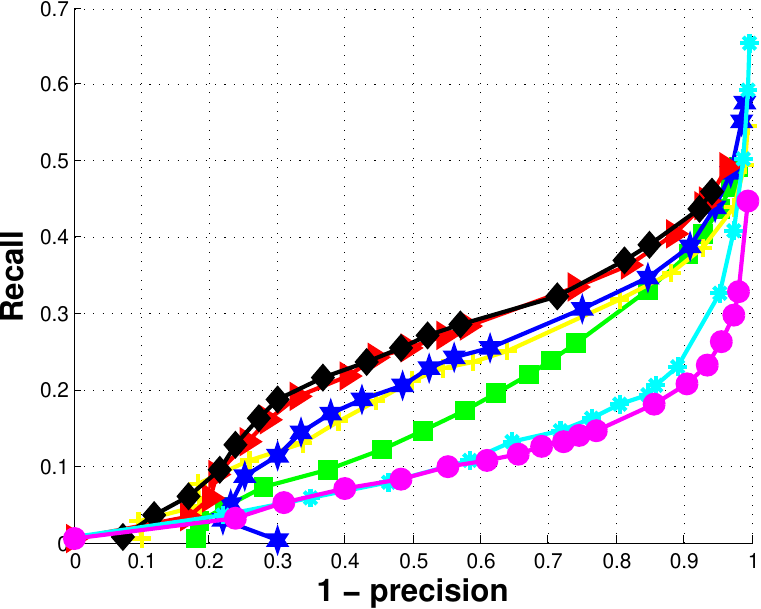}&
\includegraphics[height=28mm, width=39mm]{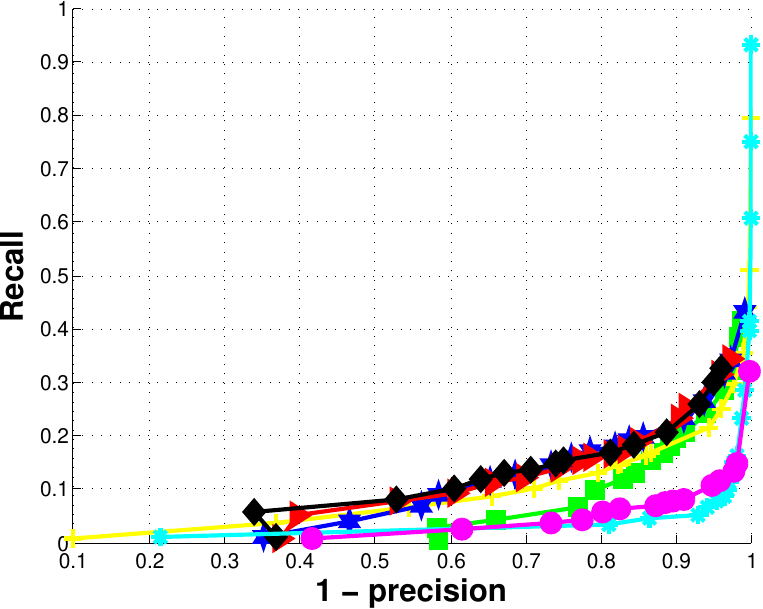}  \vspace{-.25cm} \\
\scriptsize{(f) graf 1-3}{\label{SIMgraph13}}&
\scriptsize{(g) graf 1-4}{\label{SIMgraph14}}&
\scriptsize{(h) graf 1-5}{\label{SIMgraph15}} \\

\includegraphics[height=28mm, width=39mm]{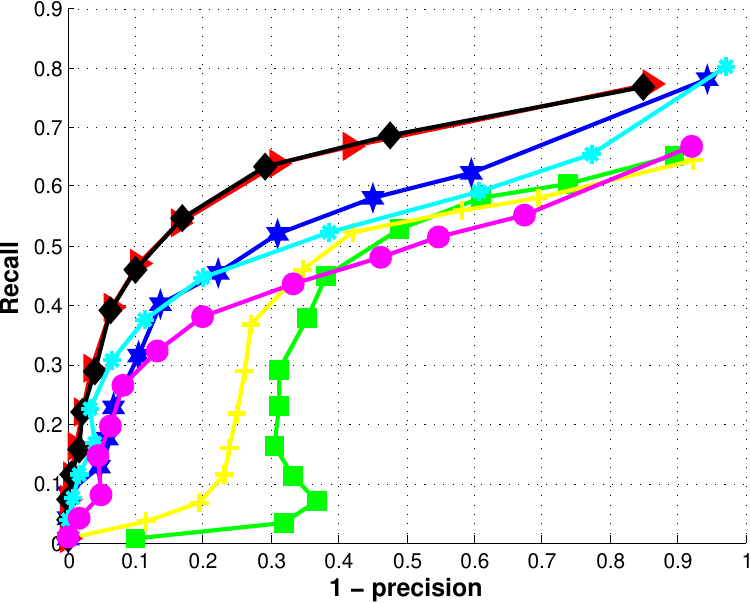}&
\includegraphics[height=28mm, width=39mm]{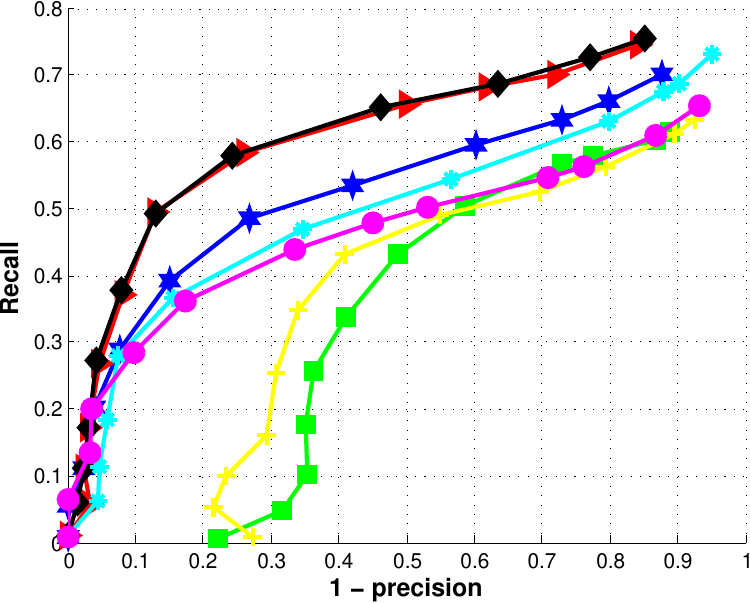}&
\includegraphics[height=28mm, width=39mm]{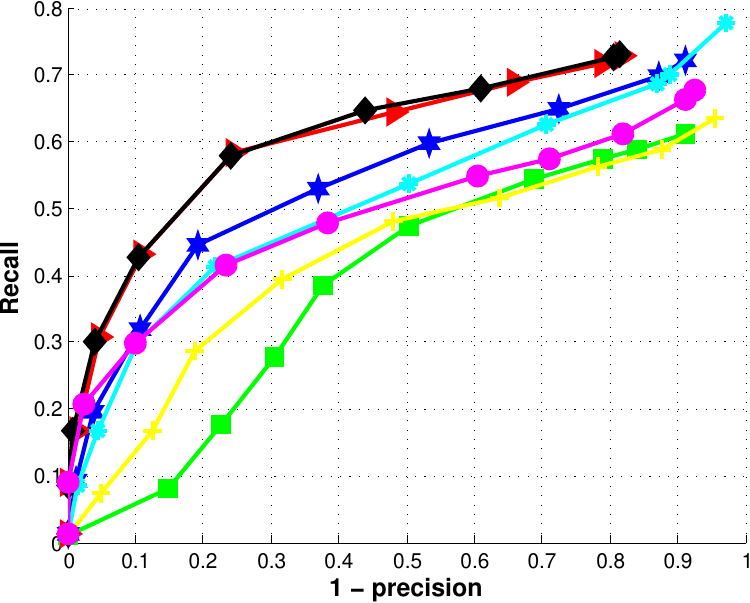} \vspace{-.25cm} \\ 
\scriptsize{(k) bike 1-3}{\label{SIMbike13}}&
\scriptsize{(k) bike 1-4}{\label{SIMbike14}}&
\scriptsize{(l) bike 1-5}{\label{SIMbike15}} \\

\includegraphics[height=28mm, width=39mm]{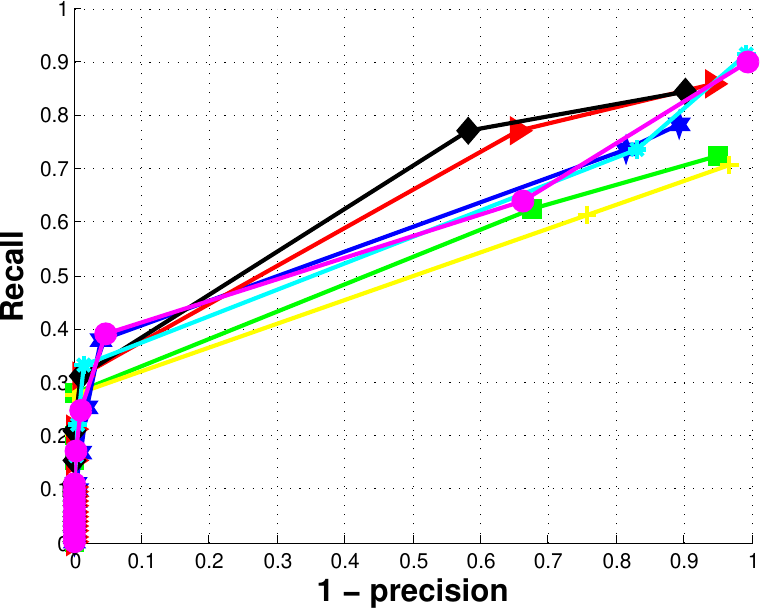}&
\includegraphics[height=28mm, width=39mm]{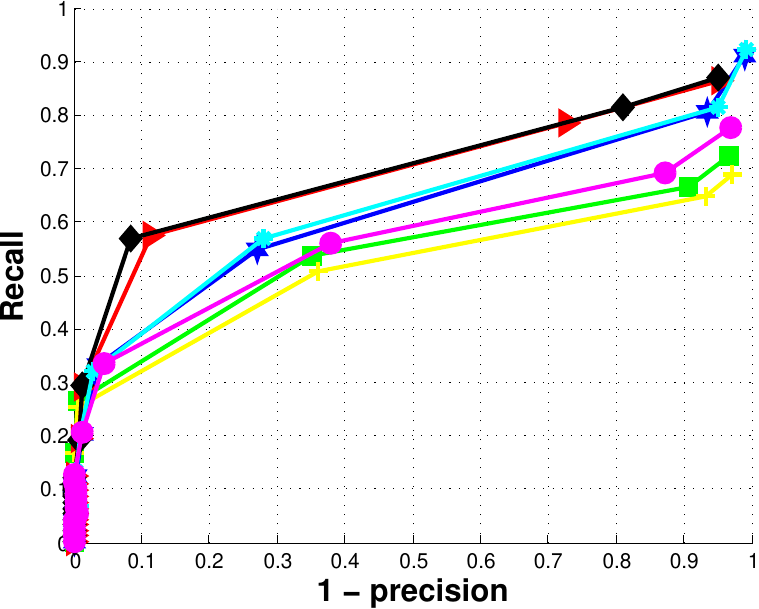}&
\includegraphics[height=28mm, width=39mm]{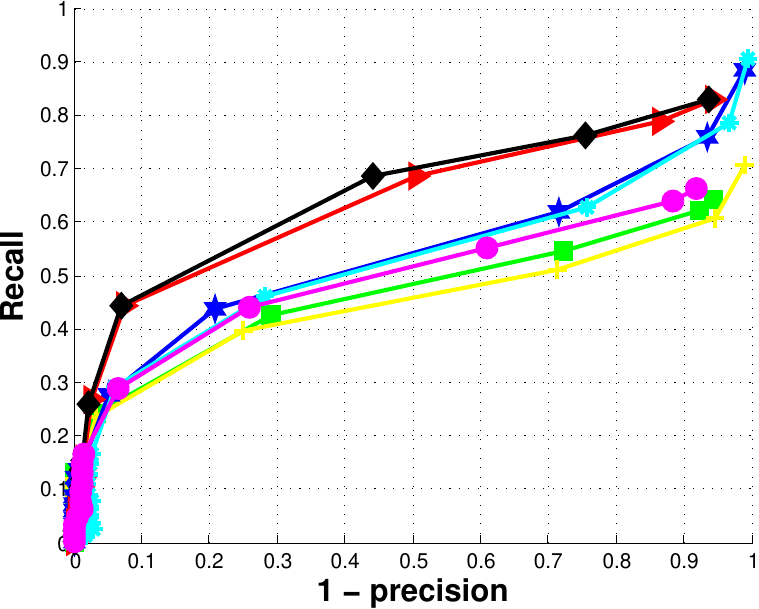}  \vspace{-.25cm} \\
\scriptsize{(n) compression 1-3}{\label{SIMcomp13}}&
\scriptsize{(o) compression 1-4}{\label{SIMcomp14}}&
\scriptsize{(p) compression 1-5}{\label{SIMcomp15}} \\

\includegraphics[height=28mm, width=39mm]{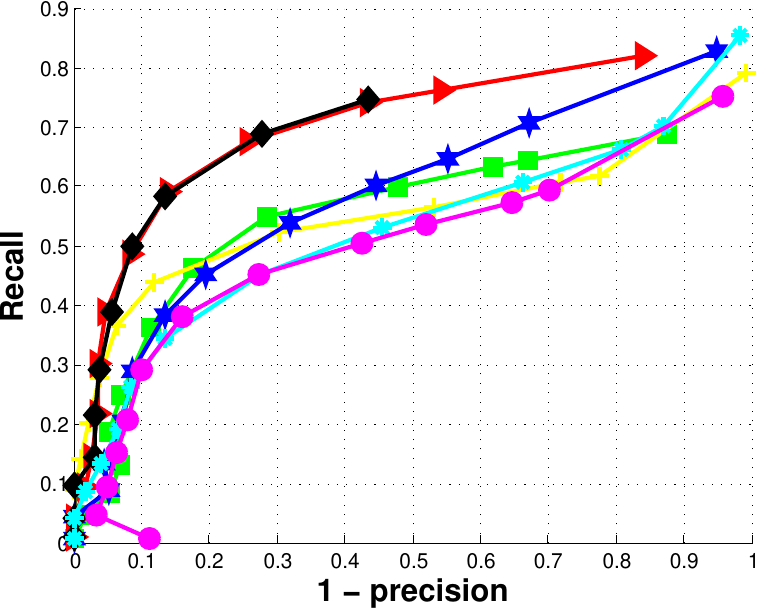}&
\includegraphics[height=28mm, width=39mm]{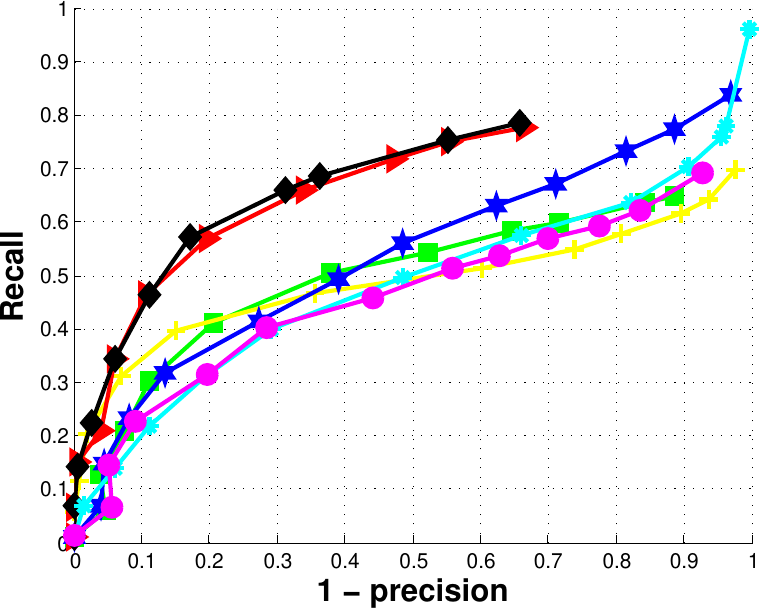}&
\includegraphics[height=28mm, width=39mm]{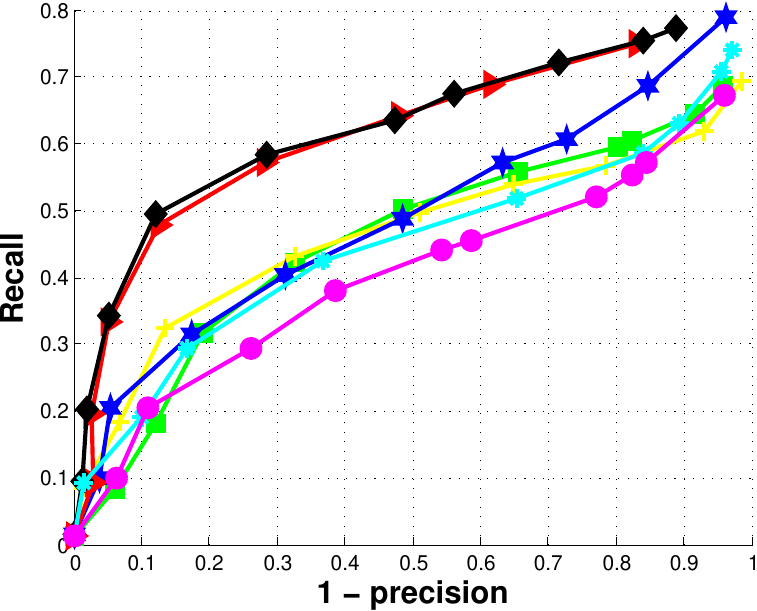}  \vspace{-.25cm} \\
\scriptsize{(r) Leuven 1-3}{\label{SIMluv13}}&
\scriptsize{(s) Leuven 1-4}{\label{SIMluv14}}&
\scriptsize{(t) Leuven 1-5}{\label{SIMluv15}} \\

\includegraphics[height=28mm, width=39mm]{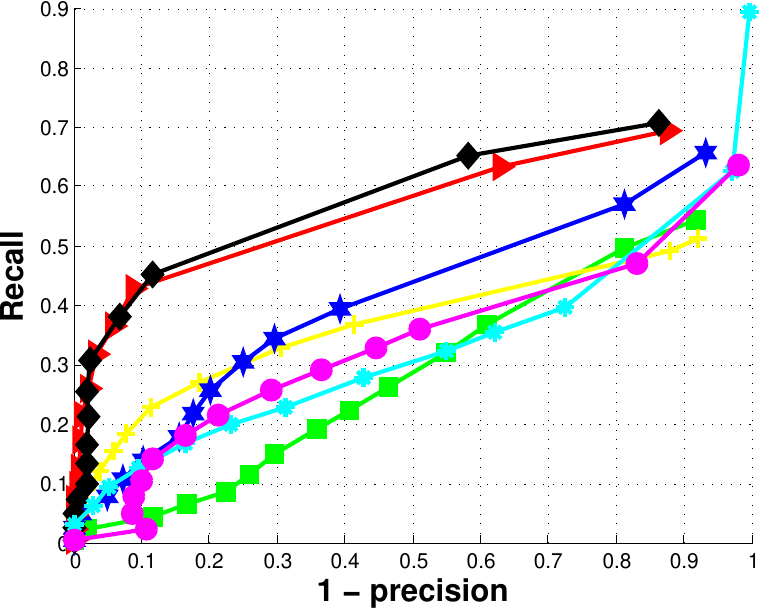}&
\includegraphics[height=28mm, width=39mm]{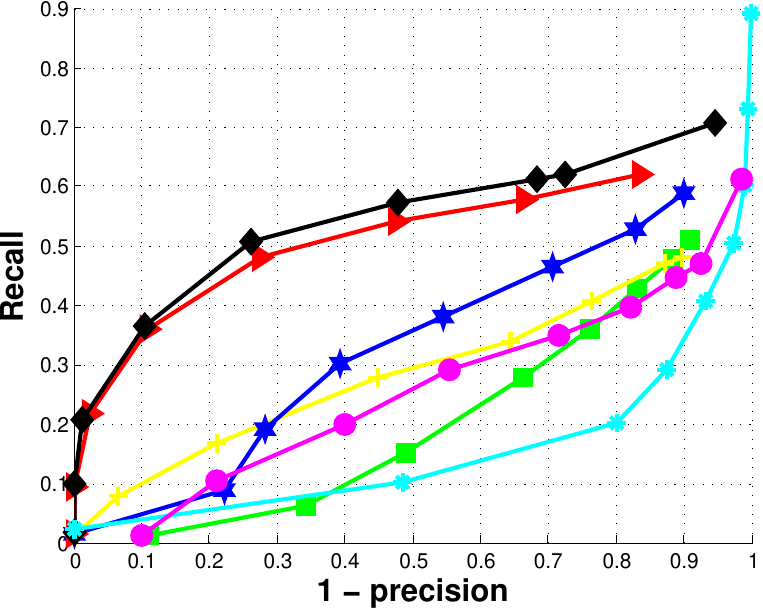}&
\includegraphics[height=28mm, width=39mm]{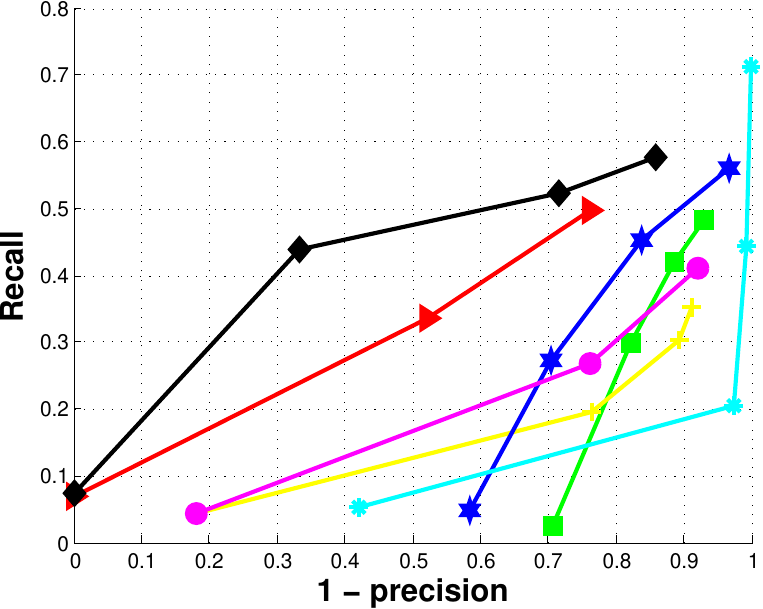}  \vspace{-.25cm} \\
\scriptsize{(r) Corridor 1-3}{\label{SIMluv13}}&
\scriptsize{(s) Corridor 1-4}{\label{SIMluv14}}&
\scriptsize{(t) Corridor 1-5}{\label{SIMluv15}} \\



\end{tabular}
\label{EVALUATION}
\end{figure}

\begin{figure}
\begin{tabular}{ccc}

\includegraphics[height=28mm, width=39mm]{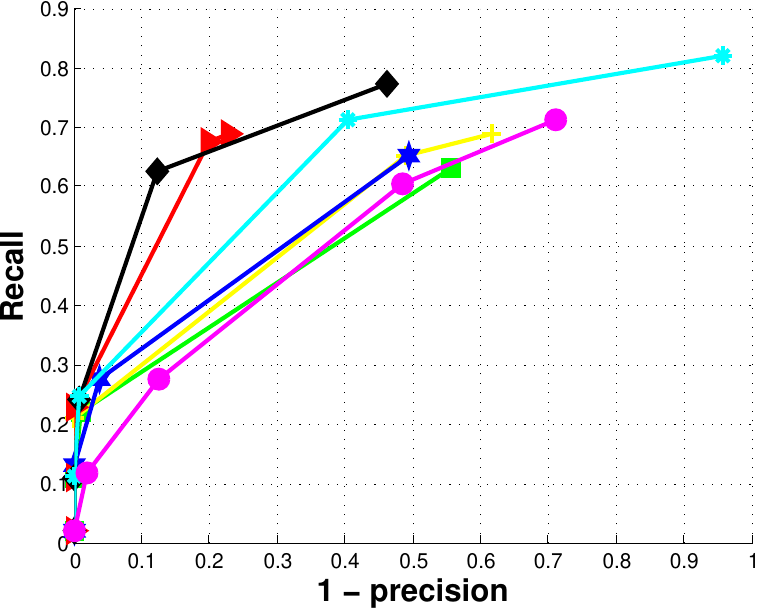}&
\includegraphics[height=28mm, width=39mm]{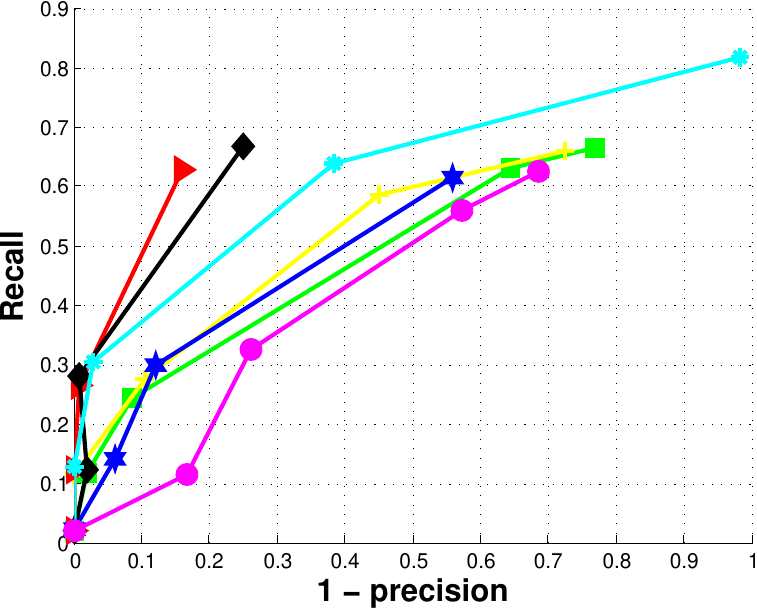}&
\includegraphics[height=28mm, width=39mm]{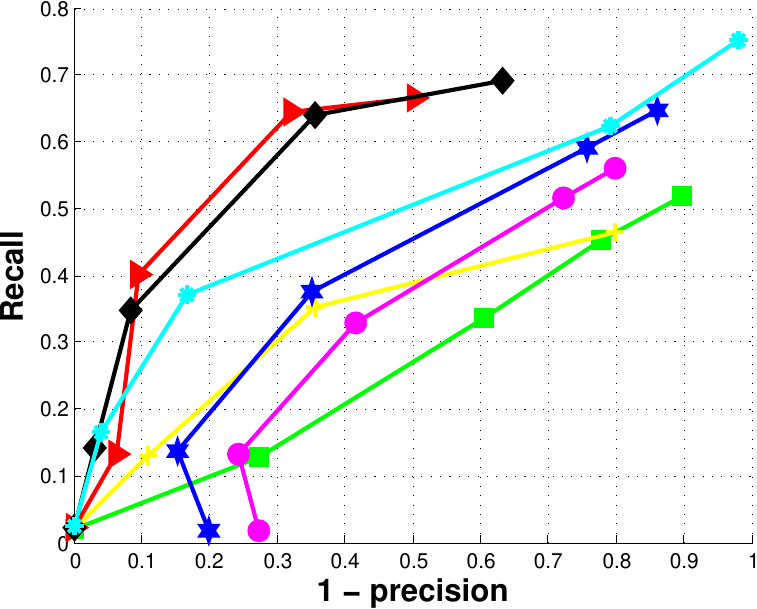}  \vspace{-.25cm} \\
\scriptsize{(r) Desktop 1-3}{\label{SIMluv13}}&
\scriptsize{(s) Desktop 1-4}{\label{SIMluv14}}&
\scriptsize{(t) Desktop 1-5}{\label{SIMluv15}} \\

\includegraphics[height=28mm, width=39mm]{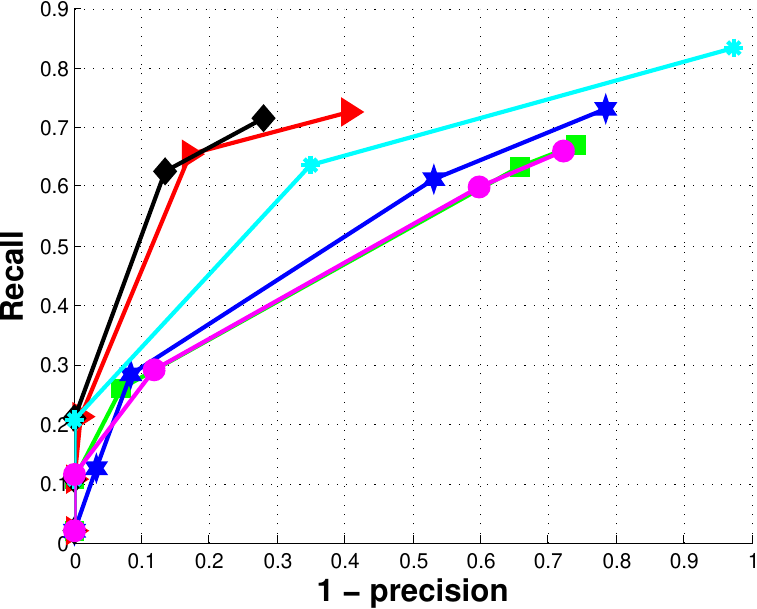}&
\includegraphics[height=28mm, width=39mm]{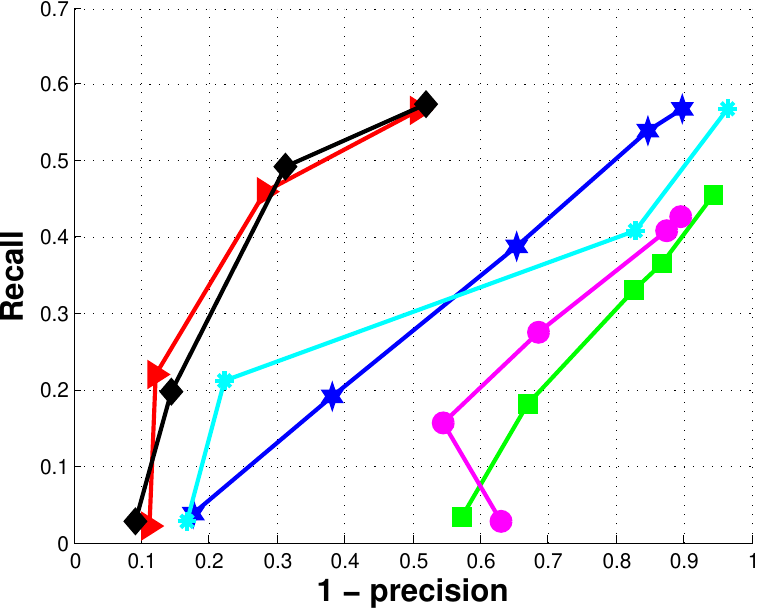}& 
\includegraphics[height=28mm , width=38mm]{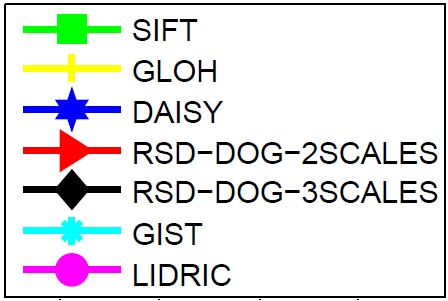} \\
\scriptsize{(s) Square}{\label{SIMluv14}}&
\scriptsize{(t) Square-root}{\label{SIMluv15}} &
\scriptsize{} \\

\end{tabular}
 \vspace{-0.15cm}
\caption{Recall vs 1-Precision curves for SIFT, GLOH, DAISY, GIST, LIRDC and RSD-DOG. Similarity matching is used for evaluation.}
\vspace{-15pt}
\label{EVALUATION}
\end{figure}

\section{Conclusion}
   The paper proposes a novel image patch descriptor based on second order statistics such as ridges, valleys, basins and so on. The originality of our method lies in combining the response of directional filter with that of the Difference of Gaussian (DOG) approach. One of the advantage of the proposed descriptor is the dimension/length. Our descriptor has a dimension of 256, which is almost 2 to 4 times less than other descriptors based on second order statistics. The experiments on complex illumination dataset illustrates the robustness of our descriptor to complex illumination changes. On the standard dataset provided by the Oxford group our descriptor outperforms SIFT, GLOH, DAISY, GIST and LIDRIC. In the future, we would like to use our descriptors for applications related to object detection, classification and image retrial. Additionally, we would like to learn the parameters by introducing a learning stage. The speed of the descriptor generation can be boosted by parallel programming.

\section{Acknowledgements}
  The work is funded by L'institut Mediterraneen des Metiers de la Longevite (I2ML).

\end{document}